\documentclass[letterpaper, 10 pt, conference]{ieeeconf}  

\IEEEoverridecommandlockouts                              

\usepackage{times}
	\usepackage{graphicx,times}
	\usepackage{subfigure}
\usepackage{amsmath}
\usepackage{amssymb}
\usepackage{mathrsfs}
\usepackage{multirow}
\usepackage{booktabs}
\usepackage{makecell}
\usepackage{tabularx}

\usepackage{algorithm, algorithmicx, algpseudocode}

\usepackage{array}
\usepackage{url}
\usepackage{mathptmx} 
\newcolumntype{P}[1]{>{\centering\arraybackslash}p{#1}}
\newcolumntype{M}[1]{>{\centering\arraybackslash}m{#1}}

\title{\LARGE \bf
Diffusion Suction Grasping with Large-Scale Parcel Dataset
}

\author{
Ding-Tao Huang, Debei Hua, Dongfang Yu, Xinyi He, En-Te Lin, Liang-hong Wang, Jin-liang Hou, Long Zeng${^*}$
\thanks{Ding-Tao Huang, En-Te Lin and Long Zeng are with the Department of Advanced Manufacturing, Shenzhen International Graduate School, Tsinghua University, Shenzhen, China (e-mail: hdt22@mails.tsinghua.edu.cn; linet22@mails.tsinghua.edu.cn; zengl@sz.tsinghua.edu.cn).}
\thanks{Xinyi He, Debei Hua and Dongfang Yu are with the Institute for Ocean Engineering, Shenzhen International Graduate School, Tsinghua University, Shenzhen, China (e-mail: hexy24@mails.tsinghua.edu.cn; hdb24@mails.tsinghua.edu.cn; ydf24@mails.tsinghua.edu.cn).}
\thanks{Liang-hong Wang and Jin-liang Hou are with the Fuwei Intelligent Technology Co., Ltd, Guangzhou, China(wanglianghon@gzfwzn.com, houjinliang@gzfwzn.com).}
\thanks{${*}$ Corresponding author.}
}
		
\begin{document}

\maketitle

\begin{abstract}

While recent advances in suction grasping have shown remarkable progress, significant challenges persist particularly in cluttered and complex parcel handling scenarios.
Current approaches are limited by (1) the lack of comprehensive parcel-specific suction grasp datasets and (2) poor adaptability to diverse object properties, including size, geometry, and texture. 
We address these challenges through two main contributions. Firstly, we introduce the Parcel-Suction-Dataset, a large-scale synthetic dataset containing 25 thousand cluttered scenes with 410 million precision-annotated suction grasp poses, generated via our novel geometric sampling algorithm. Secondly, we propose Diffusion-Suction, a framework that innovatively reformulates suction grasp prediction as a conditional generation task using denoising diffusion probabilistic models.
Our method iteratively refines random noise into suction grasping score through visual-conditioned guidance from point cloud observations, effectively learning spatial point-wise affordances from our synthetic dataset. Extensive experiments demonstrate that the simple yet efficient Diffusion-Suction achieves new state-of-the-art performance compared to previous models on both Parcel-Suction-Dataset and the public SuctionNet-1Billion benchmark. 
This work provides a robust foundation for advancing automated parcel handling systems in real-world applications.

\end{abstract}

\section{Introduction} \label{introduction}
With the growing popularity of the e-commerce industry, logistics companies are facing an increasing burden in processing and shipping the rapidly increasing volume of parcels \cite{han2020visual}. At present, traditional manual processing are no longer capable of handling such a surge in parcel volume \cite{vismanis2023robotic}. However, due to the disorderly stacked and diverse types of express packages, processing and shipping parcels present significant technical challenges. In particular, accurate and rapid object grasping has become a fundamental issue in the effective implementation of vision-based intelligent parcels processing systems.

Due to its simplicity and reliability, suction grasping is widely used in real-world tasks across various industrial and daily scenes, among the different grasping techniques \cite{Suctionnet}. Despite the success of recent object suction grasping methods, challenges still exist and need to be addressed especially in the context of the cluttered and complex parcel scenes. Two major challenges persist in obtaining robust and accurate suction grasping for parcel objects. The first challenge lies in the absence of a universal and efficient suction grasping datasets for parcel scenes. Zeng et al. \cite{zeng2022robotic} solely relied on manual experience to annotate the suctionable area and previous works \cite{CoAS-Net} proposed a synthetic suction dataset collection pipeline for cluttered environment. However, these datasets lack a general focus on the characteristics of parcel objects and a substantial amount of flat object data features. Consequently, models trained on such datasets cannot directly generalize to complex parcel scenes. Moreover, most grasp dataset samples cannot adequately account for the characteristics of large-area and severe-obstruction of objects, leading to failed suction grasping in real-world scenes.

\begin{figure}[t]
	\centering
		\includegraphics[width=1.0\columnwidth]{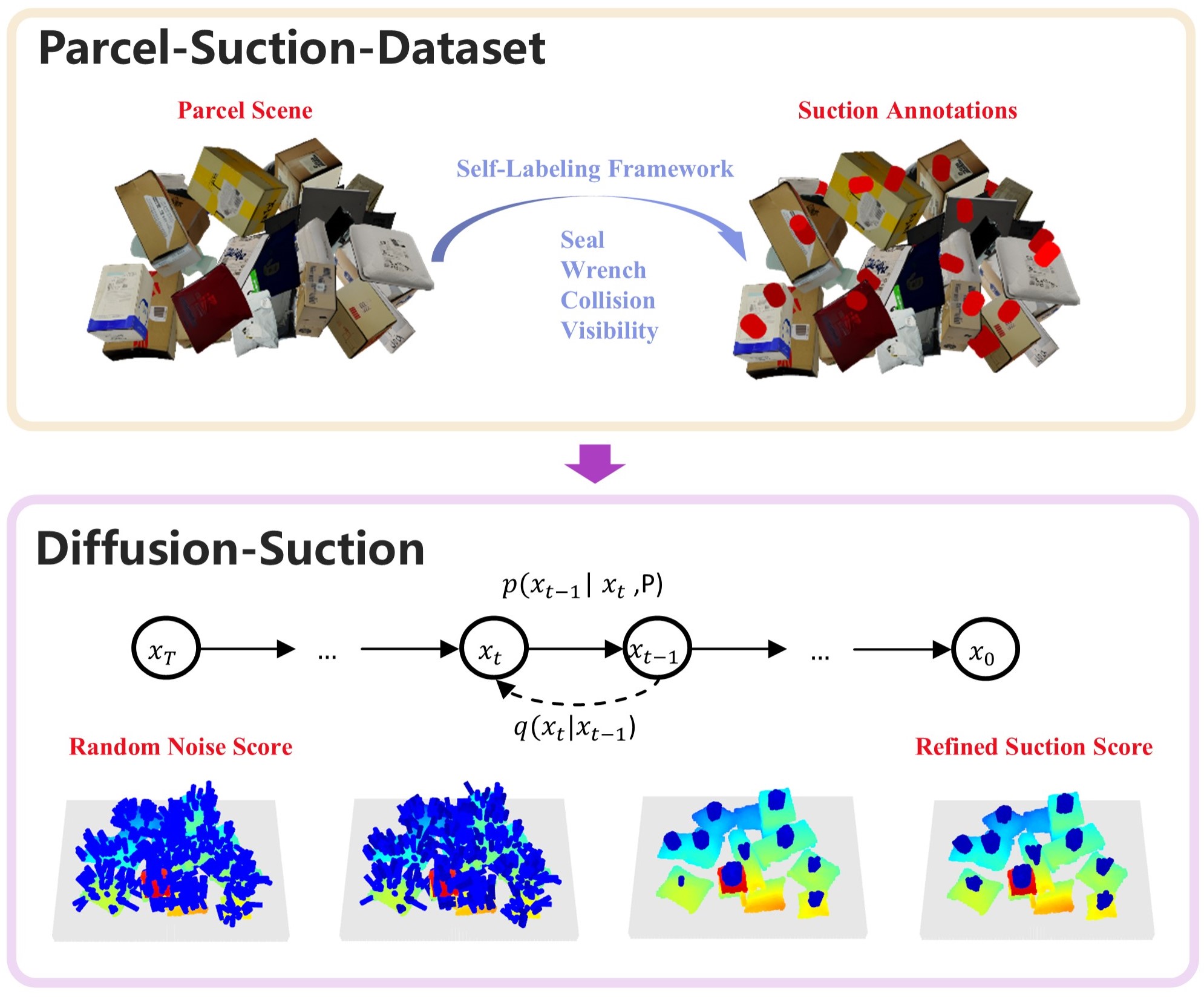}
	\caption{This work addresses two major challenges in the suction grasping task. We propose a novel pipeline that evaluates suction grasping from different perspectives to obtain annotation labels. We propose a novel framework to predict suction grasping poses by reformulating the task as an iterative diffusion-denoising process. }
	\label{fig:fig1}
\end{figure}

Secondly, suction grasping prediction is a challenging task 
because it involves finding feasible suction grasping poses given the wide range of object sizes, shapes, and textures. Many object 6D pose estimation models \cite{hernandez2017team,pvn3d,SD-Net} project the pre-defined suction configurations onto the scene. However, these approaches lack generalization capabilities for new objects. Alternatively, discriminative approaches \cite{zeng2022robotic,uncertainty,CoAS-Net} directly generate suction grasping poses from point clouds or RGB-D images. Nonetheless, these approaches are often limited by specialized network architectures and diverse training strategies to generate reliable predictions.

In this study, we introduce the Parcel-Suction-Dataset and Diffusion-Suction for suction grasping prediction in cluttered scenes, as shown in Fig. \ref{fig:fig1}. We propose a novel Self-Parcel-Suction-Labeling framework, which can automatically generate general and diverse dataset information for parcel-stacked scenes. To the best of our knowledge, this comprehensive dataset is the first large-scale synthetic suction grasping dataset specially designed for parcel stacked scenes, named Parcel-Suction-Dataset. Specifically, we propose an image to 3D parcel asset pipeline that can comprehensively capture both geometry and appearance information to obtain high-quality 3D assets. Moreover, we utilize Bullet and Blender platforms to create random unstructured parcel scenes and generate realistic synthetic rendering information, such as RGB images, segmentation images, and depth maps. We introduce a suction grasping candidate evaluation algorithm that calculates suction cup gripper's seal score, wrench score, collision score and visibility score without
requiring labor-intensive processes.

Moreover, we propose a novel Diffusion-Suction framework to predict feasible and accurate suction grasping poses by reformulating the problem as an iterative diffusion-denoising refinement process guided by 3D visual conditions. Diffusion-Suction is decoupled into the point cloud encoder and suction grasping score decoder. It utilizes the advanced point cloud feature extraction network to extract global information as condition guidance. Subsequently, we propose a novel and lightweight Pointcloud Conditioned Denoising Block, which conducts attention on the hierarchical channel and spatial information to emphasize the significant features. The point cloud encoder is executed once and the diffusion process is performed using a lightweight decoder head. With this simple and efficient structure, Diffusion-Suction significantly reduces the  computational overhead during inference. During the training stage, Gaussian noise controlled by a variance schedule \cite{ddim} is iteratively added to the ground truth to obtain noisy maps. At the inference stage, Diffusion-Suction generates reliable suction grasping scores by reversing the learned diffusion process. To the best of our knowledge, this is the first work introducing the diffusion model into suction grasping prediction task.

We evaluate our proposed method using the Parcel-Suction-Dataset and the public SuctionNet-1Billion \cite{Suctionnet} benchmark. Our experimental results demonstrate that Diffusion-Suction significantly outperforms state-of-the-art methods. We present an extensive ablation study to further reveal the effectiveness and properties of Diffusion-Suction on the Parcel-Suction-Dataset. In summary, the main contributions of our work are as follows:

\begin{itemize}
    \item We propose a novel Self-Parcel-Suction-Labeling framework to generate general and diverse large-scale Parcel-Suction-Dataset for parcel-stacked scenes.
    
    \item We formulate the suction grasping prediction task as a generative denoising process, which is the first study to apply the diffusion model to this task.

    \item We propose a novel Diffusion-Suction framework to predict feasible and accurate suction grasping poses guided by  3D visual conditions.

\end{itemize}

\section{RELATED WORK} \label{introduction}

\subsection{Suction Dataset} 
Zeng et al. \cite{zeng2022robotic} conducted a manual annotated dataset containing cluttered scenes. However, it relies solely on manual experience for annotating the suctionable and non-suctionable areas, making the annotation process highly time-consuming and expensive. Shao et al. \cite{shao2019suction} conducted a self-supervised learning robotic bin-picking system to generate training data and completed training within a few hours. However, it has only been validated on cylindrical objects, limiting the application of this method. Suctionnet \cite{Suctionnet} conducted a real word dataset which utilizes a new physical model to analytically evaluate seal and wrench formation. The annotation error depends on the accuracy of pose estimation in real scenes. Sim-Suction \cite{Sim-Suction} proposed a large-scale synthetic dataset for cluttered environments, which uses object-aware suction grasp sampling. However, there is currently a lack of readily available data in the parcel fields to explore and improve suction grasping. We propose a Self-Parcel-Suction-Labeling framework to build a large-scale parcel suction dataset which includes a variety of parcel objects collected from the real world. Furthermore, our dataset is specifically designed for parcel scenes, featuring densely stacked packages and a significant amount of flat objects.

\begin{figure*}[t]
	\centering
		\includegraphics[width=1.95\columnwidth]{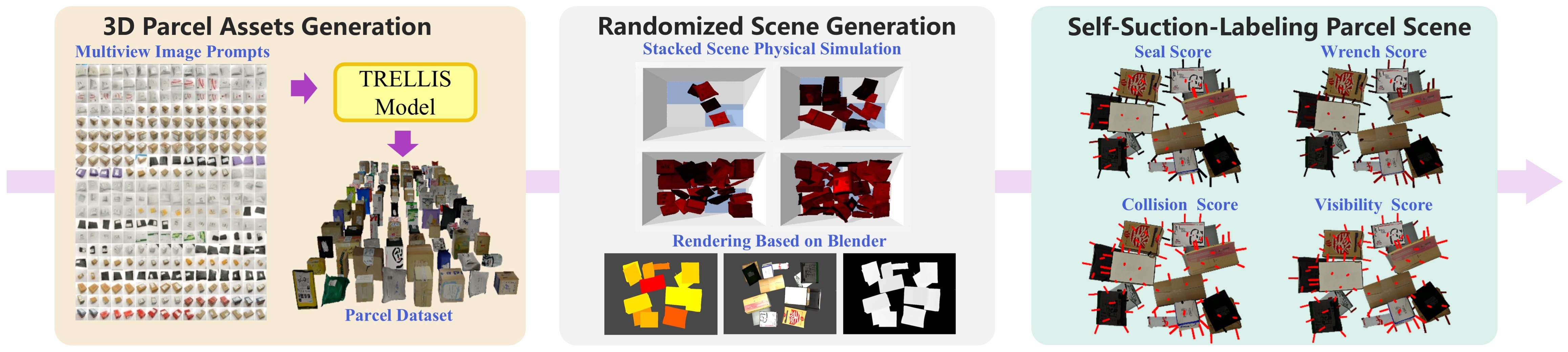}
	\caption{Overview of Self-Parcel-Suction-Labeling pipeline. Firstly, utilize the image prompts to generate 3D parcel asset model to generate high-quality 3D assets with geometry and appearance information. Next, create random unstructured parcel scenes with the Bullet and Blender simulator platform. Finally, evaluate candidate suction grasps from four different perspectives to obtain accurate annotation labels.
    }
	\label{fig:fig102}
\end{figure*}

\subsection{Suction-based Grasping} 
Hernandez et al. \cite{hernandez2017team} estimated the 6D pose of the object \cite{pvn3d,p1,p2,parametricnet,SD-Net} and projected the pre-defined suction configuration onto the objects in the scene. However, such methods lack generalization for new objects. Zeng et al. \cite{zeng2022robotic} designed a single FCN architecture trained on a human-labeled dataset, learning pixel-wise suction affordances. Suctionnet \cite{Suctionnet} proposed a pixel-wise suction scores prediction network which predicts seal score heatmap and center score heatmap separately. Sim-Suction \cite{Sim-Suction} proposed a object-aware affordance network which directly inputs text prompts to obtain identify regions of interest grasp object and outputs point-wise suction probability. Zhang et al. \cite{zhang2023robot} proposed the suction reliability matrix and suction region prediction model. However, these methods have not been tested for their effectiveness in multi-object scenes. CoAS-Net \cite{CoAS-Net} proposed a context aware suction network trained on a synthetic dataset. Rui Cao et al. \cite{uncertainty} proposed an uncertainty-aware and multi-stage framework which exploit both aleatoric and epistemic uncertainties. These methods essentially belong to discriminative approaches, while generative approaches have not yet been explored in terms of suction grasping task. Diffusion models \cite{ho2020denoising} demonstrate remarkable performance in text-to-image generation task and surpass previous generative models, such as Generative Adversarial Networks (GAN) \cite{gan} or VAE \cite{vae}. We propose to utilize the diffusion model to generate a suction grasping score map from random noise with the guidance of the visual information of the input point cloud, instead of adopting it as a normal regression head. To the best of our knowledge, this is the first work introducing the diffusion model into the suction grasping prediction task.


\section{Large-Scale Parcel Suction Dataset}

As mentioned earlier, the parcel scene has not been widely covered in existing datasets. Therefore, we propose a Self-Parcel-Suction-Labeling (SPSL) framework incorporating diverse information and construct the Parcel-Suction-Dataset as shown in Fig. \ref{fig:fig102}. This section will provide more detailed information about the SPSL.

\subsection{Randomized Scene Generation}
Acquiring high-quality 3D assets is the first step in generating scenes for dataset construction. However, there is currently a lack of available 3D parcel assets. The 3D parcel assets have the characteristics of privacy, complexity, and diversity. Using traditional laser scanning instruments to obtain 3D assets through scanning cannot effectively solve the above problems and is time-consuming. To address this issue, we propose an efficient and effective pipeline that converts images to 3D parcel assets. Specifically, we collect multiview images in real parcel stacked scenes as prompts for the latest 3D generation method TRELLIS \cite{trellis} which can comprehensively capture both structural (geometry) and textural (appearance) information. We have generated a total of 113 high-quality parcel assets and clustered them into three categories according to the Chamfer Distance and Normal Consistency metric: rectangular, planar, and cylindrical.

Using the previously generated assets, we create random unstructured parcel scenes. In the real world, parcels of various shapes and sizes are stacked randomly and disorderly into a pile. To simulate this real-world situation, we utilize the Bullet simulator platform, which is capable of simulating real dynamical collisions. Parcels are then randomly dropped into a bin with arbitrary poses on this platform. Furthermore, the Blender platform is employed to generate realistic and accurate synthetic rendering information, including RGB images, segmentation images, and depth maps. To reduce the domain gap between the synthetic and real domains, we randomize several crucial aspects of the simulation.

\begin{itemize} 
	\item Randomly sample the number of objects in the scene and the sampling range is [1,50]. 
	\item Randomly sample the pose of objects to ensure it can fall into the bin.
        \item Randomly sample the coulomb friction coefficient to model tangential forces between contact surfaces. 
        \item We apply gaussian noise augmentations with random intensity to the rendering RGB images. 
\end{itemize}

\subsection{Suction Pose Annotation}
At this stage, we will describe how to compute the suction grasping score, and the higher the score is, the easier it will be to perform suction grasping in real-world scenes. We assign four continuous scores for each suction grasping situation, namely seal score, wrench score, collision score and visibility score. We define the final metric as the product of the sub-evaluation scores $S = {S_{seal}} \times {S_{wrench}} \times {S_{collision}}\times {S_{visibility}} $.

\textbf{Seal Score Evaluation}.
The seal score is used to determine whether the suction cup can maintain a vacuum state when suction grasping is performed in a specific pose. The higher the surface flatness of the object, the higher the seal score. Similar to Cao et al \cite{Suctionnet}, we leverage the compliant suction contact model \cite{zeng2022robotic} to evaluate this process, which models the suction cup as a quasi-static spring system. Specifically, we sample dense vertices on the suction cup surface, denoted as $ \{ {v_{1}},{v_{2}},...,{v_{n}}\}$. Then we transform the suction cup model to the object surface along a specific approach vector to obtain object projection surface vertices $ \{ \widetilde {{v_1}},\widetilde {{v_2}},...,\widetilde {{v_n}}\}$. We use the distance variation between sampling points to evaluate the seal score, which can be defined as:
\begin{align}
    {S_{seal}} = 1 - max \{ {r_1},{r_2},...,{r_n}\} 
\end{align}
where the \(i\)-th sampling point distance variation \({r_i} = \min \left(1,\frac{{\widetilde {{l_i}} - {l_i}}}{{{l_i}}}\right)\). Here, \(l_i\) is the distance between two adjacent sampling points \(v_i\) and \(v_{i + 1}\) on the surface of the suction cup before the transformation of the suction cup model to the object surface. \(\widetilde{l_i}\) represents the corresponding distance between the projected points of $\widetilde {{v_i}}$ and $\widetilde {{v_{i + 1}}}$ on the object surface after the transformation. The weight of objects is not considered during seal score evaluation process. Note that we control the sampling points based on the size changes of the object, in order to achieve an optimal balance between accuracy and efficiency.

\textbf{Wrench Score Evaluation}.
The wrench score is used to determine whether the suction cup fails to resist the wrench caused by gravity in a specific pose. DexNet 3.0 \cite{zeng2022robotic} has 5 different basic wrenches criteria, which can be defined as: 
\begin{align}
    \begin{array}{l}
Friction:\sqrt 3 \left| {{f_x}} \right| \le \mu {f_N}{\rm{   }}   \quad     \sqrt 3 \left| {{f_y}} \right| \le \mu {f_N}{\rm{   }} \quad   \sqrt 3 \left| {{\tau _z}} \right| \le r\mu {f_N}{\rm{,}}\\
Material:{\rm{ }}\sqrt 2 \left| {{\tau _x}} \right| \le \pi rk{\rm{   }}  \quad   \sqrt 2 \left| {{\tau _y}} \right| \le \pi rk,\\
Suction:{f_z} \ge  - {V_f}
\end{array}
\end{align}
where $\mu $ is the friction constant, ${V_f}$ is suction grasping force, ${f_f} = ({f_x},{f_y})$ is frictional force, ${{\tau _z}}$ is torsional friction, ${\tau _e} = ({\tau _x},{\tau _{y}})$ is elastic restoring torque, $r$ is the radius suction cup and $k$ is a material-related constant. Similar to Cao et al. \cite{Suctionnet}, we have simplified these conditions by retaining only the material condition to evaluate this process. In the real world, suction grasping objects from the side often leads to failure, and humans are accustomed to suction grasping objects from the top. However, when suction grasping these objects from the side, the distance from the suction grasping point to the center of mass of the object is short. As a result, this evaluation algorithm obtains a high wrench score. Therefore, we propose a correction strategy that uses the angle between the suction grasping normal vector and gravity to correct the wrench score. Based on previous assumption, the wrench score can be defined as:
\begin{align}
{S_{wrench}} = (1 - min (1,\left| {{\tau _e}} )\right/{\tau _{thr}})(1 - a/\pi )
\end{align}
where the torque constant threshold ${\tau _{thr}} = rk\pi $ and $a$ is the angle between the suction grasping normal vector and gravity.

\textbf{Collision Score Evaluation}.
The collision score is used to determine whether the suction cup collides with objects in a specific pose in the scene. We check whether any points in the scene appear inside the mesh of the suction cup gripper. If there are points detected within the gripper, ${S_{{\rm{collision}}}}$ is set to 0, and vice versa. To balance between computational cost and accuracy, we set up a workspace around the suction cup gripper for collision detection.

\textbf{Visibility Score Evaluation}.
The visibility score is used to quantitatively reflect the degree of obstruction of the object in the scene. In highly occluded parcel scenes, some parcels are severely occluded and we are not interested in these parcel instances which cannot be captured at the bottom. Moreover, when these parcels are grabbed, it will seriously affect other objects in the scene, causing grabbing failure. Specifically, given an occluded scene image where parcels are stacked randomly into a pile, we calculate the pixel area size of each parcel, denoted as $\widetilde {{p_o}}$. Then, we iteratively remove other objects in the scene and calculate the unobstructed pixel area size of each object ${p_o}$. Visibility score can be defined as ${S_{{\rm{visibility}}}} = \widetilde {{p_o}}/{p_o}$.

\subsection{Dataset Details}
To the best of our knowledge, this comprehensive dataset is the first large-scale synthetic suction grasping dataset in parcel stacked scenes, called Parcel-Suction-Dataset. Specifically, it contains 500 cycles and each cycle contains 50 scenes. There are a total of 25 thousand scenes included. To cover the entire scene, 16384 extracted annotations are sampled from each scene using farthest point sampling. Each scene includes RGB images, segmentation images, depth maps, scene point clouds, 6D object poses, 6D object bounding boxes and camera matrices. The training set contains 100 objects, and the test set contains 13 other objects that does not appear in the training set. Notably, this Self-Parcel-Suction-Labeling framework is intended to serve as a synthetic benchmark for parcel suction grasping.

\section{Diffusion-Suction Model}

\begin{figure*}[t]
	\centering
		\includegraphics[width=1.95\columnwidth]{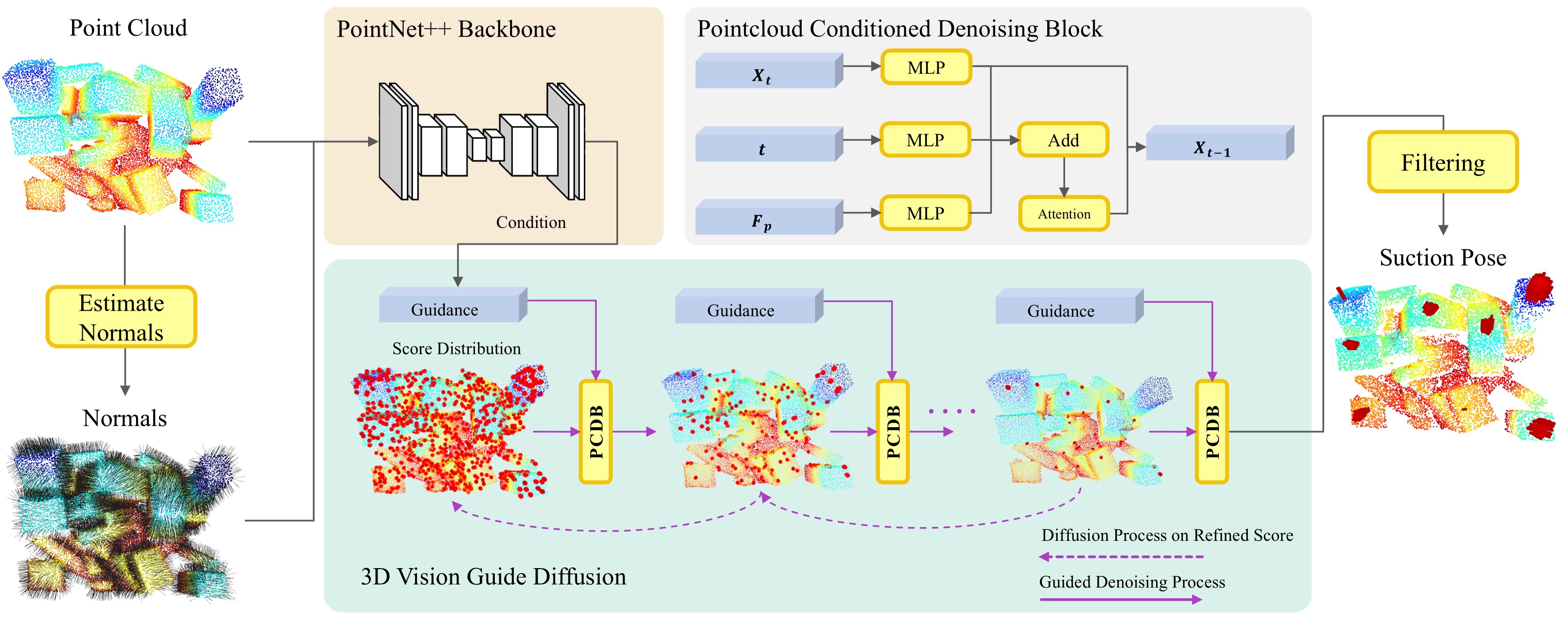}
	\caption{Overview of the Diffusion-Suction architecture. Diffusion-Suction learns an iterative denoising process to denoise random noise into suction grasping score map with the guidance of input point clouds. It uses PointNet++ to extract features as condition guidance. It consists of a novel and lightweight Pointcloud Conditioned Denoising Block. During the filtering stage, we apply Non-Maximal Suppression as a post-processing step. The red spheres represent the predicted scores ($S > 0.8$). It can be observed that the reverse process is a gradual shift from random noise to suction grasping scores.}
	\label{fig:fig103}
\end{figure*}

\subsection{Task Reformulation}

\textbf{Preliminaries}. Diffusion models \cite{ho2020denoising,song2019generative} are a class of generative models parametrized by a Markov chain, which can gradually transform a simple distribution to a structured complex distribution. These models involve two processes. The first is the diffusion process $q$, which can be defined as:

\begin{align}
	q({x_t}|{x_0})= N({x_t}|\sqrt {{{\overline a }_t}} {x_0},(1 - {\overline a _t})I)
\end{align}
which iteratively transforms desired distribution $x_0$ to a latent noisy sample $x_t$ for $t \in \{ 0,1,...,T\}$ steps. The constant ${\overline a _t} = \prod\nolimits_{s = 0}^t {{a_s}}  = \prod\nolimits_{s = 0}^t {(1 - {\beta _s})}$. ${\beta _s}$ is a constant that defines the 
added noise schedule \cite{ho2020denoising}.  

In the denoising process that reconstructs data from pure noise, the reverse process model ${\mu _\theta }({x_t},t)$ is trained to interactively predict ${x_{t - 1}}$ from ${x_{t}}$, which can be defined as:
\begin{align}
	{p_\theta }({x_{t - 1}}|{x_t}) = N({x_{t - 1}};{\mu _\theta }({x_t},t),\widetilde {{\beta _t}}I)
\end{align}
where $\widetilde {{\beta _t}} = \frac{{1 - {{\overline a }_{t - 1}}}}{{1 - {{\overline a }_t}}}{\beta _t}$ denotes the transition variance.

\textbf{Denoising as Suction Grasping}.
Given a stacked scene where parcels are stacked randomly into a pile, we are interested in detecting the suction grasping poses. The poses are defined as:
\begin{align}
	V = [t,n]
\end{align}
where $t$ represents the suction cup position on the object surface, and $n$ represents the orientation of the suction cup. We define suction grasping prediction task as follows: given the point clouds of the parcel-stacked scenes $P$, the algorithm predicts a set of feasible suction grasping poses $V$. The orientation of the suction grasping $n$ usually follows the direction of the normal vector on the object surface, and the normal vector can be calculated directly. Therefore, the suction grasping task is actually transformed into finding a set of feasible suction grasping poses $V$ with a high confidence score $x$ in the scene. Subsequently, 
the diffusion process will occur in the confidence score space. In this paper, we reformulate the suction grasping prediction as a 3D visual-condition guided denoising process, which can be defined as:
\begin{align}
    {p_\theta }({x_{t - 1}}|{x_t},P) = N({x_{t - 1}};{\mu _\theta }({x_t},t,P),\widetilde {{\beta _t}}I)
\end{align}
where the reverse process model ${\mu _\theta }({x_t},t,P)$ is trained to refine suction grasping scores from latent ${x_t}$ to ${x_{t - 1}}$ conditioned on the scene point cloud ${P}$.

\subsection{Architecture}

The diffusion model generates data samples progressively, requiring the model ${\mu _\theta }({x_t},t,P)$ to run multiple times during the inference stage. However, directly applying the model ${\mu _\theta }({x_t},t,P)$ on the raw scene point clouds at each iterative step significantly increases the computational overhead. To resolve this problem, we propose to separate the entire model into two parts: point cloud encoder and suction grasping decoder, as shown in Fig. \ref{fig:fig103}. The point clouds encoder runs only once to extract the deep feature representations from the input scene point clouds ${P}$. Then the suction grasping decoder takes these deep features as visual-condition guidance, instead of the raw point clouds ${P}$, to progressively refine the suction grasping prediction score from the noisy ${x_t}$. With this simple and efficient structure, Diffusion-Suction reduces the computational overhead during inference.

\textbf{Point cloud encoder}. It takes the original point clouds of the parcel stacked scenes as input and applies a feedforward network for feature extraction. For this purpose, we employ PointNet++\cite{pointnet++} serving as the backbone and there are other alternative backbones, e.g. DGCNN \cite{dgcnn}, Pointnet \cite{pointnet} and MinkowskiEngine \cite{Minkowski}. The extracted point-wise guidance $F_{p}$ has a size of $N_p\times N_f$. Calculating object surface normals can effectively extract the main 3D structure information of the object and provide effective prior information without a training process. Therefore, we use Open3D \cite{open3d} to directly estimate normals and then fuse the 3D normals prior feature with the point clouds to obtain accurate predictions.

\textbf{Suction grasping decoder}. The neural network model ${\mu _\theta }({x_t},t,P)$ takes available 3D visual information features from parcel stacked scene and iteratively refines suction grasping noisy map ${x_t}$. We introduce a lightweight Pointcloud Conditioned Denoising Block (PCDB) to achieve this process, which efficiently reuses shared parameters during the multi-step reverse diffusion process. Specifically, as shown in Fig. \ref{fig:fig103}, suction grasping noisy map ${x_t}$, time steps embedding $t$ and visual-condition guided $F_{p}$ are sent to different MLPs for powerful feature representation. Afterwards, the three feature maps are combined via element-wise addition. To emphasize the significant features, we conduct attention on the channel and spatial information with the convolutional block attention module (CBAM) \cite{CBAM}. Inspired by the success of the residual module \cite{he2016deep}, we conduct residual connection between the fused attention feature maps ${x_t}$ to obtain ${x_{t - 1}}$. Our PCDB balances effectiveness and efficiency, refining suction grasping with few parameters.

\subsection{Training and Inference}
We add Gaussian noise to the ground-truth suction grasping score according to Equation (5). The noise scale is controlled by ${a _t}$, which adopts the monotonically cosine schedule for ${a _t}$ in different time steps. The trainable parameters primarily include the PCDB and point cloud feature extractors. The model is trained by minimizing the loss between the diffusion suction grasping prediction and ground-truth, which is provided in Algorithm \ref{algorithm1}. The iteration number $1 \le t \le T$ is sampled from a uniform distribution, and the epsilon $\boldsymbol{\epsilon}$ is sampled from a standard distribution. Notably, the scaling factor scale controls the signal-to-noise ratio (SNR) \cite{snr}, which has a significant effect on the performance of the model. Therefore, we normalize and scale the range of the ground-truth. We found that Diffusion-Suction with a 0.5 signal scaling factor achieves optimal performance because larger scaling factors preserve a higher number of easier cases within the same time step.

\begin{algorithm}
    \caption{Diffusion-Suction Training Algorithm}
    
    \begin{algorithmic}[1]
      \Require total diffusion steps T, the scene point clouds dataset $D = \{ {P_k}\} _k^K$  
      \State  Sample $\mathbf{x}_0 \sim D$
      \State  ${F_p} = Pointnet++({{x}_0})$
      \State  Sample $\boldsymbol{\epsilon} \sim \mathcal{N}(\mathbf{0}, \mathbf{I})$
      \State  Sample $t \sim Uniform(\{1,\dots,T\})$ 
      \State  ${\overline x _0} = (sigmoid({x_0})*2 - 1)*scale $
      \State  $ loss = {\left\| {{x}_0  - {\mu _\theta }(\sqrt {{{\overline a }_t}} {{\overline x }_0} + \sqrt {1 - {{\overline a }_t}} \varepsilon ,t,{F_p})} \right\|^2}$
      
    \end{algorithmic}
    \label{algorithm1}    
\end{algorithm}

\begin{figure}[t]
	\centering
		\includegraphics[width=1.0\columnwidth]{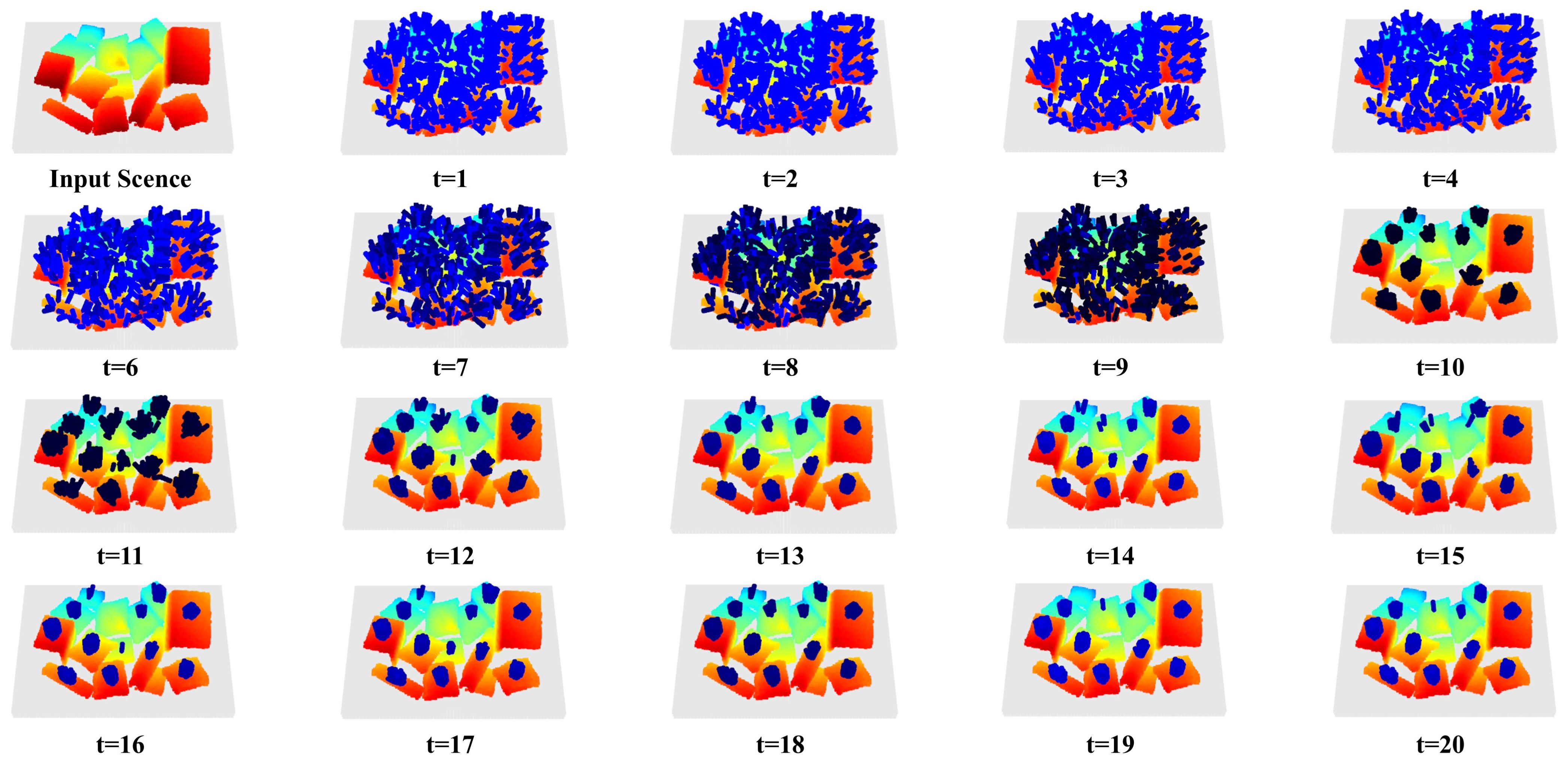}
	\caption{The figure shows the denoising reverse process with 20 inference steps. A blue cylinder represents a suction pose and the color intensity indicates the confidence of suction pose. It can be observed that the reverse process is a gradual shift from pure noise to refined suction score.}
	\label{fig:fig8}
\end{figure}

The inference procedure of Diffusion-Suction is a multi-step denoising process from a Gaussian distribution. Given raw scene point clouds as the condition input, the model progressively refines its predictions, beginning with a noise map, as shown in Algorithm \ref{algorithm2}. Moreover, we apply the DDIM \cite{ddim} inversion process to improve the inference process. Fig. \ref{fig:fig8} illustrates the denoising reverse process with 20 inference steps of Diffusion-Suction.

\begin{algorithm}
    \caption{Diffusion-Suction Sampling Algorithm}
    
    \begin{algorithmic}[1]
        \State  ${x}_t \sim \mathcal{N}(\mathbf{0}, \mathbf{I})$
        \State  ${F_p} = Pointnet++({{x}_0})$
        
        \For{$t = T,\dots,1$}
          \State ${x_{pre}} = {\mu _\theta }({x_t},t,{F_p})$
          \State ${x_t} = ddim({x_t},{x_{pred}},t,t - 1)$
        \EndFor
        
        \State \Return ${x_{pre}} $
    \end{algorithmic}
    \label{algorithm2}
\end{algorithm}

\section{Experiments}

\begin{figure*}[t]
	\centering
		\includegraphics[width=2\columnwidth]{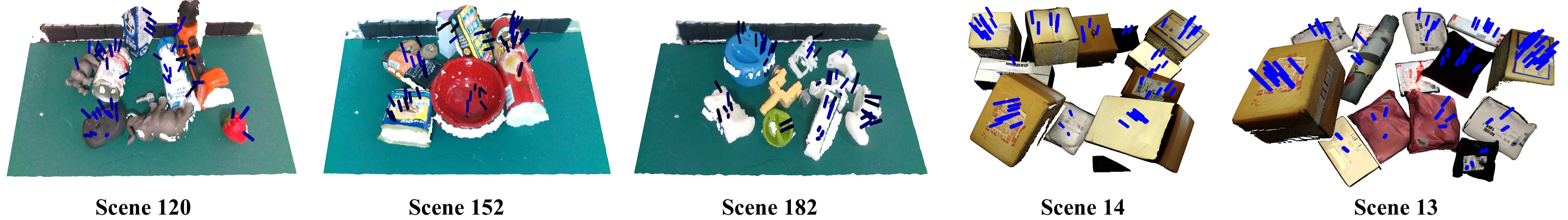}
	\caption{Qualitative results on SuctionNet-1Billion and Parcel-Suction-Dataset. Good grasp pose is marked in blue. Unsuitable pose is displayed in black.}
	\label{fig:fig9}
\end{figure*}

\begin{table*}[htb]
    \centering
    \caption{Quantitative evaluation on SuctionNet-1Billion. Evaluation results include Kinect/RealSense collection sources, where data before/after the slash correspond respectively. }
    \renewcommand{\arraystretch}{1.2} 
    \setlength{\tabcolsep}{4pt}       
    \begin{tabularx}{\textwidth}{c|c|*{3}{>{\centering\arraybackslash}X|}*{3}{>{\centering\arraybackslash}X|}*{2}{>{\centering\arraybackslash}X|}{>{\centering\arraybackslash}X}}
        \hline
        \multirow{2}{*}{Metric} & \multirow{2}{*}{Methods} & \multicolumn{3}{c|}{Seen} & \multicolumn{3}{c|}{Similar} & \multicolumn{3}{c}{Novel} \\ \cline{3-11}
                                &                          & ${AP}$ & ${AP_{0.8}}$ & ${AP_{0.4}}$ & ${AP}$ & ${AP_{0.8}}$ & ${AP_{0.4}}$ & ${AP}$ & ${AP_{0.8}}$ & ${AP_{0.4}}$ \\ \hline
        \multirow{4}{*}{Top-50} & Normal STD               & 6.16/10.10 & 0.74/1.01 & 7.20/12.64 & 17.54/19.23 & 2.38/2.26 & 24.68/25.95 & 1.34/3.50 & 0.12/0.23 & 1.08/4.22 \\ 
                                & DexNet3.0 \cite{mahler2018dex}                & 9.46/15.50 & 0.96/1.53 & 11.90/20.22 & 14.93/18.92 & 2.45/2.62 & 19.05/24.51 & 1.68/2.62 & 0.04/0.35 & 1.89/5.32 \\ 
                                & Cao et al. \cite{Suctionnet}                & 15.94/28.31 & 1.25/3.41 & 20.89/38.56 & 18.06/26.64 & 1.96/3.42 & 24.00/35.34 & 3.20/8.23 & 0.25/0.35 & 3.78/10.29 \\ \cline{2-11}
                                & Diffusion-Suction        & \textbf{22.03}/\textbf{31.90} & \textbf{3.25}/\textbf{4.79} & \textbf{30.13}/\textbf{42.78} & \textbf{19.17}/\textbf{31.59} & \textbf{2.62}/\textbf{4.27} & \textbf{25.56}/\textbf{41.38} & \textbf{4.64}/\textbf{10.00} & \textbf{0.50}/\textbf{0.76} & \textbf{5.73}/\textbf{12.32} \\ \hline
        \multirow{4}{*}{Top-1}  & Normal STD               & 11.68/15.40 & 1.94/1.82 & 13.35/18.83 & \textbf{34.11}/31.20 & \textbf{6.51}/4.74 & \textbf{47.28}/43.46 & 2.64/4.89 & 0.28/0.52 & 3.72/5.96 \\ 
                                & DexNet3.0 \cite{mahler2018dex}              & 12.30/10.61 & 3.36/1.30 & 15.03/13.91 & 20.75/14.28 & 5.39/2.40 & 26.00/17.71 & 1.09/2.81 & 0.08/0.29 & 1.17/3.15 \\ 
                                & Cao et al. \cite{Suctionnet}              & 26.44/43.98 & 2.90/7.79 & 35.98/60.81 & 24.59/36.59 & 2.45/6.54 & 32.89/49.30 & 6.66/\textbf{14.32} & 0.13/0.38 & \textbf{8.80}/\textbf{19.95} \\ \cline{2-11}
                                & Diffusion-Suction        & \textbf{35.97}/\textbf{54.79} & \textbf{13.68}/\textbf{21.03} & \textbf{47.55}/\textbf{68.59} & 19.31/\textbf{47.00} & 5.18/\textbf{16.08} & 25.36/\textbf{59.36} & \textbf{7.49}/14.14 & \textbf{2.88}/\textbf{2.42} & 8.46/17.67 \\ \hline
    \end{tabularx}
    \label{tab:table100}
\end{table*}

\begin{table}[!t]
	\centering
		\caption{Quantitative evaluation on the Parcel-Suction-Dataset. }

        \begin{tabular}{c|c|ccc}
\hline
Metric                  & Methods &${AP}$ & ${AP_{0.8}}$ & ${AP_{0.4}}$ \\ \hline
\multirow{4}{*}{Top-50} & Normal STD      & 50.98  & 6.09   & 73.16   \\ \cline{2-2}
                        & DexNet3.0 \cite{mahler2018dex}         & 9.56  & 0.46   & 9.47    \\ \cline{2-2}
                        & Cao et al. \cite{Suctionnet}       &62.81    & 28.29   &  78.66  \\ \cline{2-5} 
                        & Diffusion-Suction       &	\textbf{94.42}	&	\textbf{85.04}	& \textbf{97.65}   \\ \hline
\multirow{4}{*}{Top-1}  & Normal STD     &  55.75   &2.00    &  84.00  \\ \cline{2-2}
                        & DexNet3.0 \cite{mahler2018dex}        &  6.50  & 0.00   &  4.00  \\ \cline{2-2}
                        & Cao et al. \cite{Suctionnet}      & 66.25   & 30.00   &  82.00  \\ \cline{2-5} 
                        & Diffusion-Suction       &	\textbf{98.00}	&	\textbf{92.00}	& \textbf{100.00}   \\ \hline
\end{tabular}
	\label{tab:table99}
\end{table}

\subsection{Datasets and evaluation metrics}
We evaluate our method on two datasets: our proposed Parcel-Suction-Dataset containing 25 thousand annotated stacked parcel scenes, and the public SuctionNet-1Billion benchmark \cite{Suctionnet} to demonstrate the framework's effectiveness.
This dataset contains 190 scenes and we collect 256 perspectives for each scene. The dataset is divided into three subsets to test performance in different scenes: seen, similar, and novel.

For performance evaluation, we adopt the Average Precision (AP) metric following the protocol in \cite{Suctionnet}, which computes seal and wrench scores for each prediction online. If the product of the seal score and the wrench score exceeds the threshold of 0.4 or 0.8, the predicted suction pose is considered correct. When evaluating the Parcel-Suction-Dataset, we calculate visibility scores for online evaluation.

\subsection{Evaluation on Parcel-Suction-Dataset}

To validate the performance of our proposed approach on the Parcel-Suction-Dataset, we compare Diffusion-Suction against several state-of-the-art approaches, as shown in Table \ref{tab:table99}. As we can observe, our proposed approach achieves state-of-the-art results. It outperforms other methods with an improvement of +32\% on average precision the in the top-50 metric. Moreover, our Diffusion-Suction achieves great improvement in the Top-1 metric, indicating its ability to generate the best suction grasping candidate in cluttered scenes, which is crucial for real-world suction grasping applications. Notably, in terms of the ${AP_{0.8}}$ metric, our method significantly outperforms other approaches, showing the most substantial improvement. This indicates that most of suction grasping poses predicted by Diffusion-Suction have high scores. Fig. \ref{fig:fig9} illustrates the qualitative results of Diffusion-Suction on the cluttered and complex parcel scenes, showing the Top-50 suction grasps.

\subsection{Evaluation on SuctionNet-1Billion}
Table \ref{tab:table100} summarizes the comparison results between our proposed approach and current suction grasping methods on public benchmark. Overall, our proposed approach significantly outperforms three state-of-the-art approaches across both RealSense and Kinect data source. Specifically, it’s important to note that our proposed approach demonstrates consistently better performance in all the top-50 benchmark compared with other methods. 
This superior performance can be attributed to our diffusion head's advanced 3D visual-condition feature extraction capability, which better captures global information and leads to more accurate predictions.
In addition, our proposed approach outperforms other models in all the seen scenes and exhibits strong learning capability for seen objects. It shows that utilizing diffusion model on this task can achieve robustness and efficiency prediction in terms of the seen scenes. Our proposed approach also demonstrates competitive performance on the similar and novel scenes.

\subsection{Ablation study}
We conduct extensive ablation studies on the Parcel-Suction-Dataset to analyze the key components and properties of Diffusion-Suction.

\textbf{Framework settings}.
We report the compatibility of the proposed model, as shown in Table \ref{tab:table104}. The Diffusion-Suction (w/o Normal) and Diffusion-Suction (w/o Visibility) results demonstrate the importance of visibility score decoder and 3D normal prior feature fusion. Moreover, this indicates that the proposed framework with PointNet++ \cite{pointnet++} for feedforward feature extraction can achieve optimal performance compared to PointNet \cite{pointnet} and DGCNN \cite{dgcnn} model.

\textbf{Denoising step}.
Training and inference step not only affect inference speed, but also require high GPU memory consumption. To explore the balance between accuracy and efficiency, we change the training and inference step, as shown in Table \ref{tab:table107}. Changing the inference step directly can result in a significant decrease in performance. However, keeping the training and inference step the same and reducing the number of step slightly decrease the performance, but it can accelerate the model speed.

\begin{table}[!t]
	\centering
		\caption{Ablation study on different Diffusion-Suction framework settings on the Parcel-Suction-Dataset. w/o, without.}
		\begin{tabular}{c|c|c|c}
		\Xhline{0.5pt}

Method	&	${AP}$	&	${AP_{0.8}}$	&	${AP_{0.4}}$		\\	\hline

Diffusion-Suction (w/o Normal)		  & 93.09	&	81.06	&	97.23			\\
Diffusion-Suction (w/o Visibility)		      & 89.15	&72.08		&	95.59			\\
Diffusion-Suction (PointNet)		  & 38.39	&	8.72	&	48.67			\\
Diffusion-Suction (DGCNN)		      & 78.41	&	39.56	&	93.76			\\
Diffusion-Suction    &	\textbf{94.42}	&	\textbf{85.04}	& \textbf{97.65}  	\\

		\Xhline{0.5pt}
		\end{tabular}
	\label{tab:table104}
\end{table}

\begin{table}[!t]
\centering
\caption{Ablation study on different training and inference step. }

\begin{tabular}{cc|c|c|c|c}
\hline
\multicolumn{2}{c|}{Method} & \multirow{2}{*}{${AP}$} & \multirow{2}{*}{${AP_{0.8}}$} & \multirow{2}{*}{${AP_{0.4}}$}  & \multirow{2}{*}{Time(ms)}\\ \cline{1-2}
Training Step  & Inference Step &                    &                    &                    \\ \hline

t = 5      & t = 5         & 94.17                 & 84.00                  & 97.61         & 181         \\ 
t = 10      & t = 10         & 94.23                 &84.12                  & 97.63       &  185         \\ 
t = 15      & t = 15         & 93.57                  & 82.36                & 97.36        & 194         \\
t = 20      & t = 5         & 28.10                  & 22.99                  & 29.86        & 181         \\
t = 20      & t = 10         & 81.95                  & 48.54                  & 95.03        &188          \\ 
t = 20      & t = 15         & 10.06                  & 9.73                  & 10.15         &193         \\  \hline
t = 20      & t = 20          &	\textbf{94.42}	&	\textbf{85.04}	& \textbf{97.65} 	& 204    \\ \hline

\end{tabular}

	\label{tab:table107}
\end{table}

\begin{table}[!t]
        \centering
        \caption{Robotic grasping experiments on the real-world parcel scenes with the proposed approach Diffusion-Suction.}

\begin{tabular}{c|cc|cc}  
\hline
\multirow{2}{*}{Object} & \multicolumn{2}{c|}{Scattered Scene} & \multicolumn{2}{c}{Stacked Scene}  \\
\cline{2-3}
\cline{4-5}
& ${R_{grasp}}$ & ${R_{object}}$  & ${R_{grasp}}$ & ${R_{object}}$  \\ \hline
Rectangular               & 96.43      & 100.0      & 93.75      & 100.0      \\ 
Planar               & 100.0      & 100.0      & 94.64      & 100.0      \\ 
Cylindrical               & 81.36      & 90.83      & 78.69      & 100.0      \\ 
Mixed               & 94.44      & 97.62      & 93.33      & 100.0          \\ \hline
\end{tabular}
\label{tab:table108}
\end{table}

\subsection{Real World Experiment}
\begin{figure}[t]
	\centering
		\includegraphics[width=1.0\columnwidth]{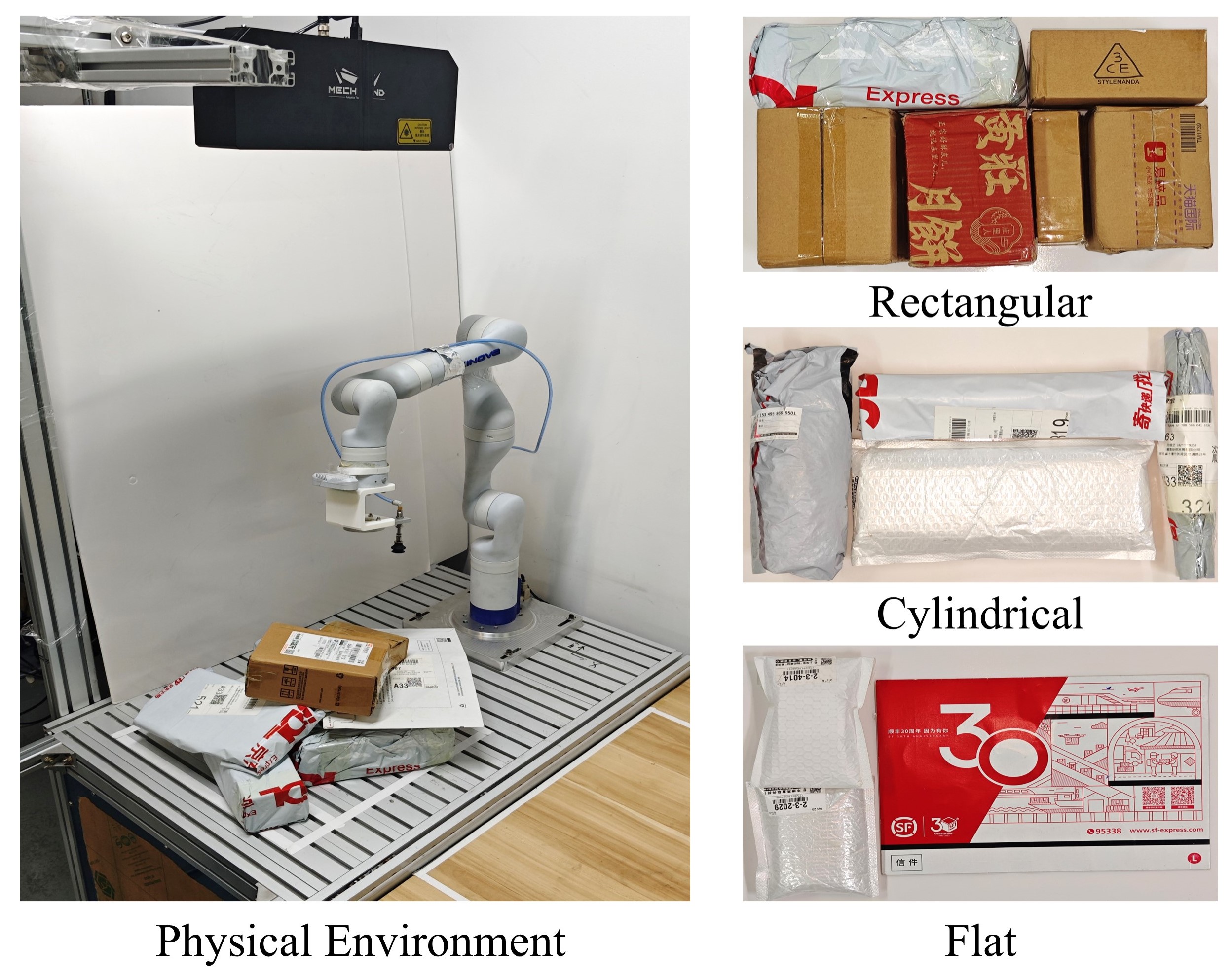}
	\caption{Real-world grasp experiment setup.}
	\label{fig:fig10}
\end{figure}

Real-robot experiments are conducted using the Kinova Gen3 robot and a Mech-Eye industrial 3D camera, as shown in Fig. \ref{fig:fig10}. To rigorously evaluate robustness across diverse grasping conditions, we conducted tests in diverse scenarios involving four types of objects: rectangular, flat, cylindrical, and mixed (a combination of the former three). Additionally, we configured scenes according to difficulty levels as scattered and stacked Scenes. Two metrics are adopted: 1) grasping success rate ${R_{grasp}}$ = Successful grasps/Total attempts, 2) object clearance rate ${R_{object}}$ = Cleared objects /Total objects. The experiment are terminated if the robot failed to grasp any object in three consecutive attempts. Quantitative results, as shown in Table \ref{tab:table108}, demonstrate our method's effectiveness in real-world parcel scenes.

\section{Conclusions}
In this paper, we introduce the Parcel-Suction-Dataset and Diffusion-Suction for suction grasping prediction in parcel cluttered scenes. The Parcel-Suction-Dataset comprises 25 thousand cluttered scenes with 410 million annotated suction grasping poses. Moreover, Diffusion-Suction denoise random prior noise into suction grasping score map with the guidance of input point clouds visual-condition. Experiments show that Diffusion-Suction improvements average precision compared to well-established models on Parcel-Suction-Dataset and public SuctionNet-1Billion benchmarks. We will release the code and dataset soon (\url{https://github.com/TAO-TAO-TAO-TAO-TAO/Diffusion_Suction}). 


\bibliographystyle{IEEEtran}
\bibliography{IEEEabrv,ParametricNet}









\end{document}